\documentclass[preprint,review,12pt]{elsarticle}

\usepackage{framed,multirow}
\usepackage{url}

\usepackage{amssymb, amsbsy, algorithmic, algorithm}
\usepackage{amsmath}

\usepackage{epsfig}
\usepackage{epstopdf} 
\usepackage{url}
\usepackage{multirow}
\usepackage{graphicx}
\usepackage{booktabs}
\usepackage{color}
\usepackage{amsbsy}
\usepackage{balance}
\usepackage{citesort} 
\usepackage{soul}
\usepackage[mathcal]{euscript}
\usepackage{glossaries}

\newcommand{\argmin}{\operatornamewithlimits{argmin}}

\usepackage{lipsum}
\makeatletter
\def\ps@pprintTitle{%
 \let\@oddhead\@empty
 \let\@evenhead\@empty
 \def\@oddfoot{}%
 \let\@evenfoot\@oddfoot}
\makeatother
\begin{document}

\begin{frontmatter}

\title{Sparse Representation-Based Classification: Orthogonal Least Squares or Orthogonal Matching Pursuit?}

\author[1]{Minshan Cui} 

\author[1]{Saurabh Prasad}

\address[1]{Department of Electrical and Computer Engineering, University of Houston \\ Contact Author: Saurabh Prasad, Email: saurabh.prasad@ieee.org}


\begin{abstract}
Spare representation of signals has received significant attention in recent years. Based on these developments, a sparse representation-based classification (SRC) has been proposed for a variety of classification and related tasks, including face recognition. Recently, a class dependent variant of SRC was proposed to overcome the limitations of SRC for remote sensing image classification. Traditionally, greedy pursuit based method such as orthogonal matching pursuit (OMP) are used for sparse coefficient recovery due to their simplicity as well as low time-complexity. However, orthogonal least square (OLS) has not yet been widely used in classifiers that exploit the sparse representation properties of data. Since OLS produces lower signal reconstruction error than OMP under similar conditions, we hypothesize that more accurate signal estimation will further improve the classification performance of classifiers that exploiting the sparsity of data. In this paper, we present a classification method based on OLS, which implements OLS in a classwise manner to perform the classification.  We also develop and present its \emph{kernelized} variant to handle nonlinearly separable data. Based on two real-world benchmarking hyperspectral datasets, we demonstrate that class dependent OLS based methods outperform several baseline methods including traditional SRC and the support vector machine classifier.
\end{abstract}


\end{frontmatter}

\section{Introduction}
In recent years, sparse representation of signals has drawn considerable interest and has shown to be powerful in many applications --- particularly in compression and denoising. It is based on the observation that most natural signals can be sparsely represented in an appropriate representation. Applications of sparse signal representations can be found in various fields such as image denoising \citep{elad2006image, dabov2007image}, restoration \citep{mairal2008sparse}, visual tracking \citep{bai2012robust,lu2013robust}, detection \citep{zhu2014sparse,cong2013abnormal}, and classification \citep{WYG2009,chen2015vehicle,zhou2013kernel,wang2012supervised}. Recent work in \citep{WYG2009}, Wright \emph{et al.} proposed a sparse representation-based classification (SRC) for face recognition. The basic idea of SRC is to learn a sparse representation for a test sample as a (sparse) linear combination of all training samples (over-complete dictionary), wherein the class-specific dictionary yielding the lowest reconstruction error determines the class label for the test sample. SRC has also been actively applied in various classification problems including vehicle classification \citep{mei2011robust}, multimodal biometrics \citep{shekhar2014joint}, digit recognition \citep{labusch2008simple}, speech recognition \citep{gemmeke2011exemplar}, hyperspectral image classification \citep{CNT2011, MStgars2013}. 

Finding the sparsest solution in SRC is a combinatorial problem as it involves searching through every combination of $S$ atoms in a dictionary, where $S$ denotes the optimal sparsity level. There are two major approaches to approximate this problem. One is to relax this non-convex combinatorial problem into an $\ell_1$ convex  optimization problem --- also known as basis pursuit. Several methods have been proposed to solve this $\ell_1$-norm problem including interior-point method \citep{KKLG2007}, gradient projection \citep{FNW2007} etc. The other major category is based on iterative greedy pursuit algorithms such as matching pursuit, orthogonal matching pursuit (OMP) and orthogonal least square (OLS). These greedy approaches have been widely used due to their computational simplicity and easy implementation. They find an atom at a time based on different criterion and update the sparse solution iteratively. 
Among these approaches, the OMP algorithm is by far the most popular approach and is used in a wide range of applications. The main difference between OMP and MP is that OMP uses an orthogonal dictionary while MP does not. Making the dictionary orthogonal will reduce the redundancy of the dictionary when estimating the signal. OLS is similar to OMP except for the atom selection process. A major difference between OMP and OLS relies on their atom selection procedure in that OMP selects an atom that best correlates with the current residual, while OLS selects an atom giving the smallest residual after orthogonalization. The time complexity of OMP is $O(dnS)$ where $d$ is number of features, $n$ is the dictionary size and $S$ is the sparsity level. The time complexity of OLS is slightly higher than OMP which is caused by the difference in the atom selection process. Note that the first atom selected by OMP is identical to OLS. For more detailed information about the differences between these two algorithms, readers can refer to  \citep{blumensath2007difference,rebollo2002optimized} and a $k$-step analysis of OMP and OLS can be found in \citep{soussen2011joint}.

OLS has been widely used in many applications \citep{chen2006local,chen2009orthogonal,huang2005determining,huang2012orthogonal,zhang2014two}, but it has not  gained much attention for classification problems. In \citep{MStgars2013}, the authors implement SRC in a classwise manner to improve the classification accuracy, in which the sparse coefficient is recovered by OMP. In this work, we implement A class-dependent version of OLS to perform classification. {Since OLS produces lower signal reconstruction error compared to OMP under similar condition \citep{blumensath2007difference} (such as the same sparsity level, same dictionary etc.) --- an observation that will be further analyzed and explained in the next section, we hypothesize that more accurate signal estimation will further improve the classification performance of SRC. Compared with convex optimization based techniques such as interior point and gradient projection methods \citep{sjk,mat}, greedy pursuit-based approaches are more efficient and appropriate to recover the sparse coefficient in SRC due to their low time-complexity. By using the kernel trick, we extend the proposed cdOLS into its kernel variant to handle nonlinearly separable data as well.}

The remainder of this paper is organized as follows. In Sec.~\ref{sec:sparse}, we briefly introduce the basic concept of SRC and illustrate the recovery performance of OMP and OLS using an illustrative case study. The proposed cdOLS as well as its kernel variant are also described in Sec.~\ref{sec:sparse}. Experimental hyperspectral datasets and comparative classification results are presented in Sec.~\ref{sec:experiment}. We provide concluding remarks in Sec.~\ref{sec:conclusion}.

\section{Sparse representation}
\label{sec:sparse}
\subsection{Sparse representation-based classification}
Assume $\boldsymbol{a}_{ij} \in \mathbb{R}^{d}$ represent the $j$-th training sample from class $i$, $\boldsymbol{A} = [\boldsymbol{A}_{1},\boldsymbol{A}_{2}, \hdots, \boldsymbol{A}_{c}]$, where $\boldsymbol{A}_{i} = [\boldsymbol{a}_{i1},\boldsymbol{a}_{i2}, \hdots, \boldsymbol{a}_{i n_{i}}] \in \mathbb{R}^{d \times n_{i}}$ is the $i$-th class training sample set, $c$ is the number of classes, $n_{i}$ represents the number of training samples from class $i$, and $n$ is the total number of training samples, $n = \sum_{i=1}^{c}{n_{i}}$. 

Based on the assumption of SRC, a test sample $\boldsymbol{x} \in \mathbb{R}^{d}$ from class $i$ approximately lies in the linear span of training samples from class $i$ which can be described as 
\begin{eqnarray}
	\boldsymbol{x} & \approx & \beta_{i1}\boldsymbol{a}_{i1} + \beta_{i2}\boldsymbol{a}_{i2} + \hdots + \beta_{in_{i}}\boldsymbol{a}_{in_{i}} \nonumber \\
	& = & [\boldsymbol{a}_{i1},\boldsymbol{a}_{i2}, \hdots, \boldsymbol{a}_{in_{i}}] [\beta_{i1},\beta_{i2}, \hdots, \beta_{in_{i}}]^{\top} \nonumber \\
	& = & \boldsymbol{A}_{i} \boldsymbol{\beta}_{i} 
\end{eqnarray}
where $\boldsymbol{\beta}_{i}$ is a coefficient vector whose entries are the weights of the corresponding training samples in $\boldsymbol{A}_{i}$.

In real-world classification problems, the true label of the test sample is unknown. Thus $\boldsymbol{x}$ needs to be represented as a linear combination of all training samples in $\boldsymbol{A}$ as described below
\begin{equation}
	\boldsymbol{x} = \boldsymbol{A}\boldsymbol{\beta}
	\label{eq:linear}
\end{equation}
where $\boldsymbol{\beta} = [\beta_{11},\beta_{12},  \hdots, \beta_{cn_{c}}]$ is a coefficient vector corresponding to $\boldsymbol{A}$. 

Ideally, the entries of $\boldsymbol{\beta}$ are all zeros except those related to the training samples from the same class as the test sample. The residual of each class can be calculated via
\begin{equation}
	\mathbf{r}_i(\boldsymbol{x}) = \|\boldsymbol{a} - \boldsymbol{A}_{i}\hat{\boldsymbol{\beta}}_{i}\|_{2}, \quad \quad i = 1,2, \ldots, c 
\end{equation}
where $\hat{\boldsymbol{\beta}}_{i}$ denotes the entries of the coefficient vector $\boldsymbol{\beta}$ associated with the training samples from the $i$-th class. 

Finally, $\boldsymbol{x}$ is assigned a class label $i$ corresponding to a class that resulted in the minimal residual.

\subsection{Sparse solution via OMP and OLS}
\label{sec:omp_ols}

The sparsest solution of $\boldsymbol{x}$ in \eqref{eq:linear} can be obtained by solving
\begin{equation}
	\hat{\boldsymbol{\beta}} = \operatorname{argmin}\|\boldsymbol{\beta}\|_{0}, \quad \operatorname{s.t.} \quad \boldsymbol{A}\boldsymbol{\beta} = \mathbf{x}, 
	\label{eq:l0}
\end{equation}
where the l0-norm $\|\cdot\|_{0}$ simply counts the number of nonzero entries in $\boldsymbol{\beta}$. 

The problem in \eqref{eq:l0} is NP-hard, and it cannot be solved in polynomial time. There are several different approaches \citep{CDS1998,KKLG2007,wipf2004sparse} to solving this sparse approximation problem in~\eqref{eq:l0}, in this letter, we focus on the two greedy pursuit based approaches --- OMP and OLS. 

Both OMP and OLS can be used to approximate the sparsest solution in~\eqref{eq:l0}. In each iteration, the atom selected by OMP is not designed to minimize the residual norm after projecting the target signal onto the selected elements, while OLS selects the atom that minimizes the residual based on the previously selected atoms. Thus the final residual norm generated by OLS is always smaller than OMP under similar conditions. However, OLS does not always give the sparsest solution. To find an optimal $S$-term representation of an signal $\boldsymbol{x}$ in~\eqref{eq:l0}, a simple approach to finding the sparsest solution then is to search over all possible linear combinations of $S$ atoms in $\boldsymbol{A}$. Let us denote this exhaustive searching algorithm as combinatorial orthogonal least square (COLS). The first atom selected by OLS and OMP is the same and fixed. However, COLS iteratively select each of the atom as the first atom and remaining atoms are selected based on OLS. Specifically, it first selects the first atom and then select the remaining ($S - 1$) atoms based on OLS. After selecting $S$ atoms, it uses them to estimate the signal and calculates the residual  (least square error) between the signal and the estimated signal. Following this, it selects the next atoms as the first set of atoms and repeats the above process. After calculating all $n$ ($n$ is the dictionary size) residuals using each atom as the first atom, it chooses the minimal residual as the final output. This is further explained graphically in fig. 1 next.

We use an intuitive example to illustrate the differences of OMP, OLS and COLS algorithms. {In \citep{blumensath2007difference}, the authors use a graphical interpretation to show the difference between OMP and OLS in terms of atom selection procedure. In this example, we will further illustrate that the norm of residual generated by OLS is smaller than OMP but they are both not optimal. We will demonstrate later that the signal reconstruction performance of OLS is close to optimal.} Assume the true sparsity level in~\eqref{eq:l0} is $S$. Let $\boldsymbol{z}_1$, $\boldsymbol{z}_2$ and $\boldsymbol{z}_3$ be the axes in a 3-dimensional space, and $\boldsymbol{a}_1$, $\boldsymbol{a}_2$, $\boldsymbol{a}_3$ be the atoms in a dictionary $D$. Without loss of generality, assume $\boldsymbol{a}_1$ and $\boldsymbol{z}_1$ are overlapped with each other, and $\boldsymbol{a}_2$ and $\boldsymbol{a}_3$ are in the $\boldsymbol{z}_1\boldsymbol{z}_2$-plane and $\boldsymbol{z}_1\boldsymbol{z}_3$-plane respectively. Let $\boldsymbol{x}$ be a target signal, and assume that $\boldsymbol{a}_1$ is the most correlated with $\boldsymbol{x}$ than $\boldsymbol{a}_2$ and $\boldsymbol{a}_3$. Let $\vec{OF} = \vec{AD}$. Let $\phi_1$ and $\phi_2$ be the angles between $\boldsymbol{a}_2$ and $\vec{OF}$, and $\boldsymbol{a}_3$ and $\vec{OF}$ respectively. Under this scenario, we will analyze the optimal sparse $S$-term representation using OMP, OLS and COLS, where $S$ equals to 2. 1) OMP first selects the most correlated atom which is $\boldsymbol{a}_1$, and produces the residual $\vec{AD}$ by projecting $\boldsymbol{x}$ onto it. Next, OMP selects an atom that is mostly correlated with $\vec{AD}$. Since $\vec{OF}=\vec{AD}$ and $\phi_1 < \phi_2$, OMP selects $\boldsymbol{a}_2$. Therefore, the final residual norm produced by OMP is $\|\vec{AB}\|_2$, which is obtained by projecting $\boldsymbol{x}$ onto $\boldsymbol{a}_1\boldsymbol{a}_2$-plane. 2) For OLS, the first atom selected is $\boldsymbol{a}_1$, since OMP and OLS are the same in the first iteration. Next, OLS calculates the residual norms of $\|\vec{AC}\|_2$ and $\|\vec{AB}\|_2$ obtained by projecting $\boldsymbol{x}$ onto $\boldsymbol{a}_1\boldsymbol{a}_3$-plane and $\boldsymbol{a}_1\boldsymbol{a}_2$-plane respectively, and selects $\boldsymbol{a}_3$, since $\|\vec{AC}\|_2 < \|\vec{AB}\|_2$. Thus, the final residual norm of OLS is $\|\vec{AC}\|_2$ obtained by projecting $\boldsymbol{x}$ onto $\boldsymbol{z}_1\boldsymbol{z}_3$-plane. 3) COLS calculates all residuals by projecting $\boldsymbol{x}$ onto  planes formed by every combination of two atoms. Since $\|\vec{AE}\|_2 < \|\vec{AC}\|_2 < \|\vec{AB}\|_2$, COLS selects $\boldsymbol{a}_2$ and $\boldsymbol{a}_3$. The final residual norm is $\|\vec{AE}\|_2$. For the special case when $D$ is an orthonormal dictionary, all of the above three methods will find an optimal $S$-term representation \citep{tropp2004greed}. Overall, the performance of these methods with regard to the reconstruction error are COLS $\ge$ OLS $\ge$ OMP. 

\begin{figure}[htbp] 
	\centering
	\includegraphics[width=7cm]{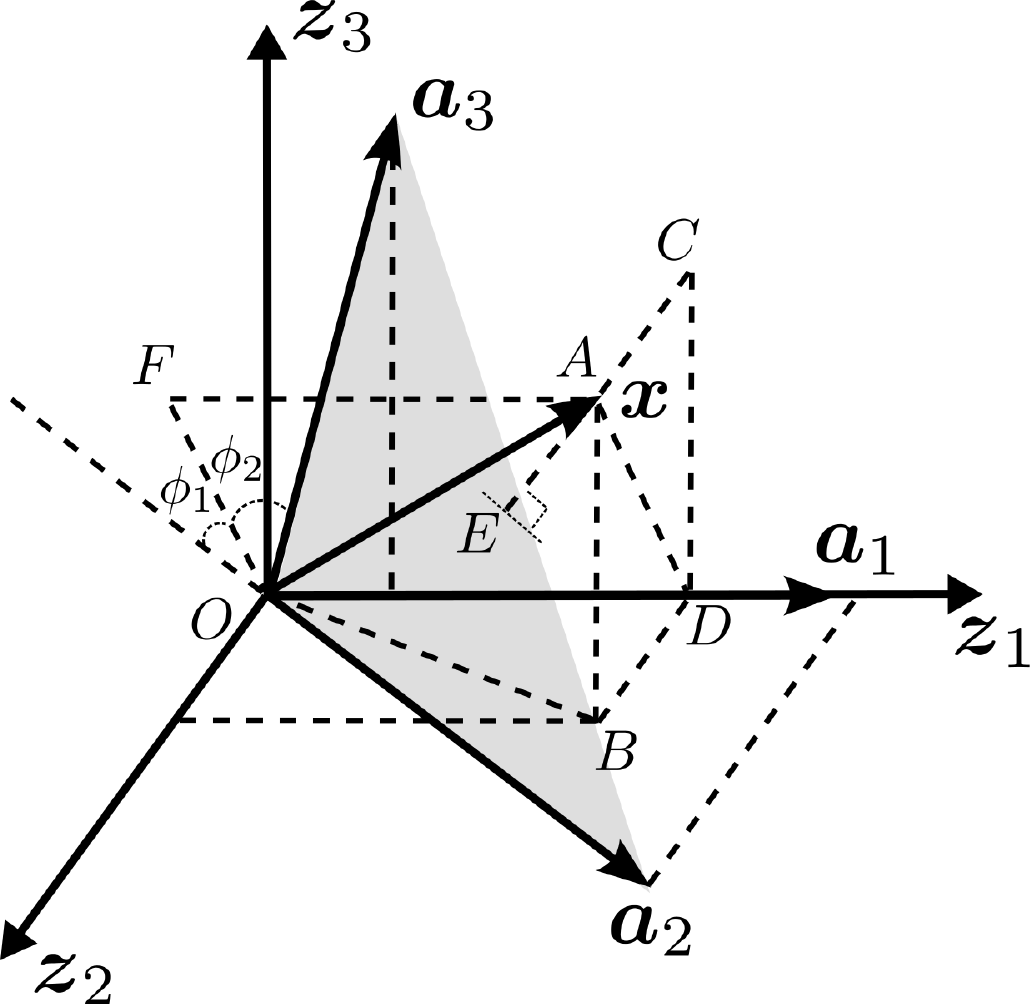} \\
	\caption{Graphically illustrating OMP, OLS and COLS.
	}
	\label{fig:omp_ols}
\end{figure}


\subsection{The proposed OLS-based classification}
\label{sec:classification}
The recent work in \citep{MStgars2013} demonstrates that operating SRC in a class-wise manner can significantly improve the classification performance of SRC. As is explained in the previous section, the recovery ability of OLS is always better than OMP in terms of the least square error under the same condition (i.e. the same sparsity level). Therefore, it is expected that the classification performance can be significantly enhanced by replacing OMP with OLS under this framework. We name this algorithm class-dependent OLS (cdOLS). Note that the stopping criterion in cdOLS is based on the sparsity level. This is because the signal estimation error monotonically decreases as the sparsity level increases. {Hence, we use the same sparsity level for each class to circumvent this bias.} We also extend cdOLS to a ``kernel'' cdOLS (KcdOLS). The cdOLS and KcdOLS algorithms are described in Algorithm~\ref{alg:cdols} and Algorithm~\ref{alg:kcdols}. For a faster implementation of OLS, readers can refer to \citep{blumensath2007difference}.

\begin{algorithm}[htbp]
	\caption{\emph{cdOLS}}
	\label{alg:cdols}
	\begin{algorithmic}[1]
		\STATE \textbf{Input:} A training dataset $\boldsymbol{A} \in \{\boldsymbol{A}_l\}_{l=1}^{c} \in \mathbb{R}^{d \times n}$, test sample $\boldsymbol{x} \in \mathbb{R}^{d}$ and sparsity level $S$.
		\\\hrulefill
		\FORALL{$l \in {1,2, \ldots, c}$}
		\STATE Set $\Lambda^{0} = \emptyset$, $\boldsymbol{r}^{0} = \boldsymbol{y}$, and iteration counter $m = 1$. 
		\WHILE{$m \le S$}
		\STATE Update the support set $\Lambda^{m} = \Lambda^{m-1} \cup \lambda^{m}$ by solving
		\begin{equation}
			\lambda^{m} = \argmin_{j=1,2,\ldots, n}\|\boldsymbol{x} - (\boldsymbol{A}_l)_{:,\Lambda^{m\!-\!1}\cup j}\tilde{\boldsymbol{\beta}}\|_2, \nonumber
		\end{equation} 
		where $\tilde{\boldsymbol{\beta}} = (\boldsymbol{A}_l^{\dagger})_{:,\Lambda^{m\!-\!1}\cup j}\boldsymbol{x}.$
		\STATE Calculate the residual $\boldsymbol{r}^{m}$ by solving 
		\begin{equation}
			\boldsymbol{r}^{m} = \boldsymbol{x} - \boldsymbol{A}_{:,\Lambda^{m}}\hat{\boldsymbol{\beta}}, \nonumber
		\end{equation}
		where $\hat{\boldsymbol{\beta}} = (\boldsymbol{A}^{\dagger}_l)_{:,\Lambda^{m}}\boldsymbol{x}$.
		\STATE $m \leftarrow m + 1$ 
		\ENDWHILE
		\STATE Calculate the $l$-th class residual norm $\displaystyle{\nu_l = \|\boldsymbol{r}^{m-1}\|_2}$.
		\ENDFOR
		\STATE Class label of $\boldsymbol{x}$: 
		$\displaystyle{\omega = \argmin_{l = 1,2,\ldots, c}\nu_l}$.
		\\\hrulefill	
		\STATE \textbf{Output:} A class label $\omega$.	
	\end{algorithmic}
\end{algorithm}

\section{Experimental Validation}
\begin{figure*}[t]
	\centering
	\includegraphics[width=15cm]{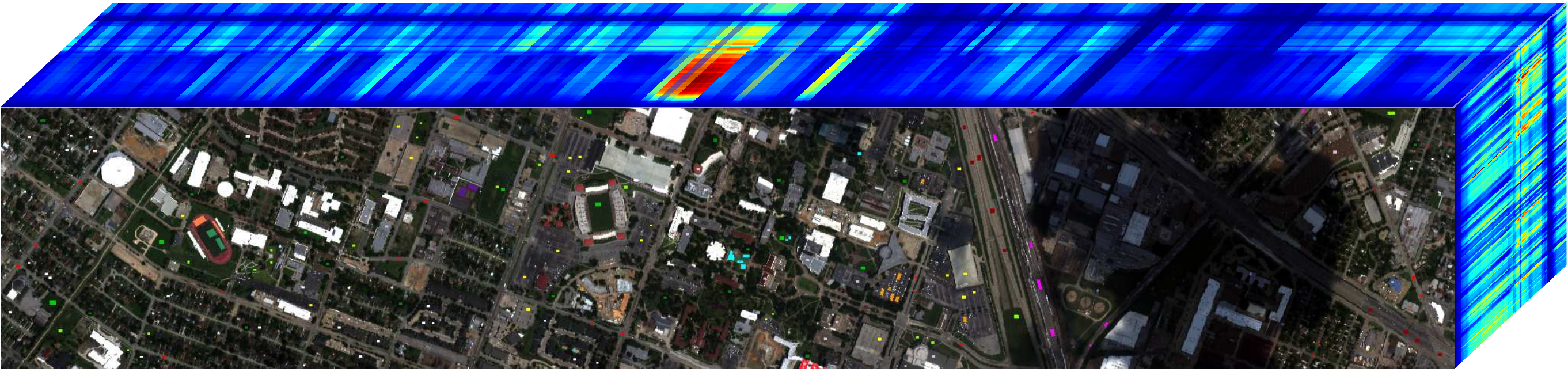} \\
	\vspace*{0.1in}
	\includegraphics[width=14cm]{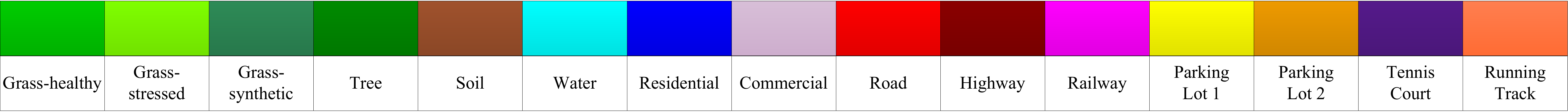}
	\caption{
		True color image inset with ground truth for University of Houston hyperspectral data.}
	\label{data_uh}
\end{figure*}

We validate the proposed cdOLS and KcdOLS and compare with various baselines using two benchmark hyperspectral datasets. The first dataset is acquired using an ITRES-CASI (Compact Airborne Spectrographic Imager) 1500 hyperspectral imager over the University of Houston campus and the neighboring urban area in 2012. This image has {a spatial dimension of $1905 \times 349$ with a} spatial resolution {of} $2.5m$. There are 15 number of classes and 144 spectral bands over {the} $380 - 1050nm$ wavelength range. Fig.~\ref{data_uh} shows the true color image of University of Houston dataset inset with the ground truth. 

\begin{figure}[htbp] 
	\centering
	\begin{tabular}{cc}
		\includegraphics[width=4cm]{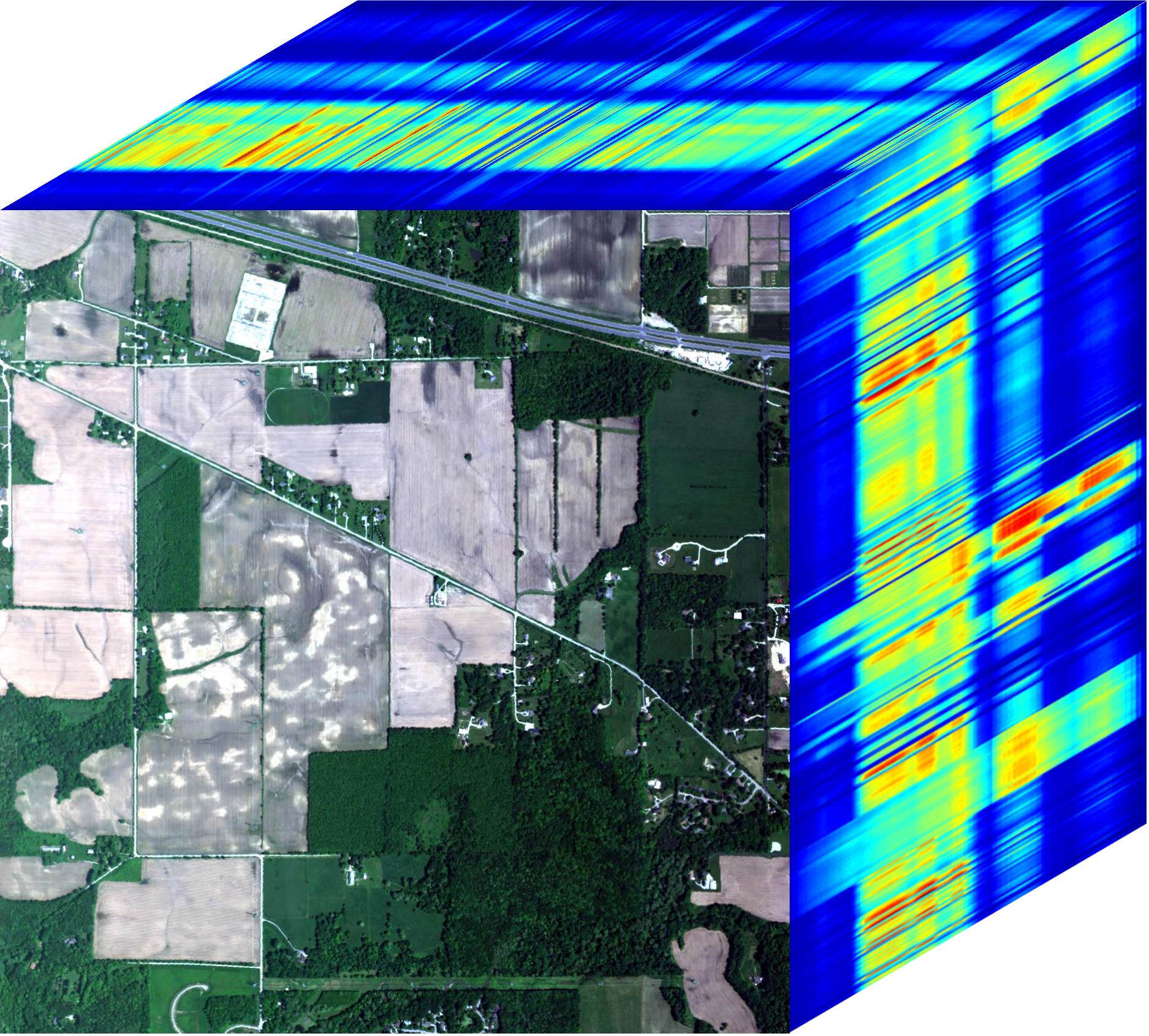} &
		\includegraphics[width=3.45cm]{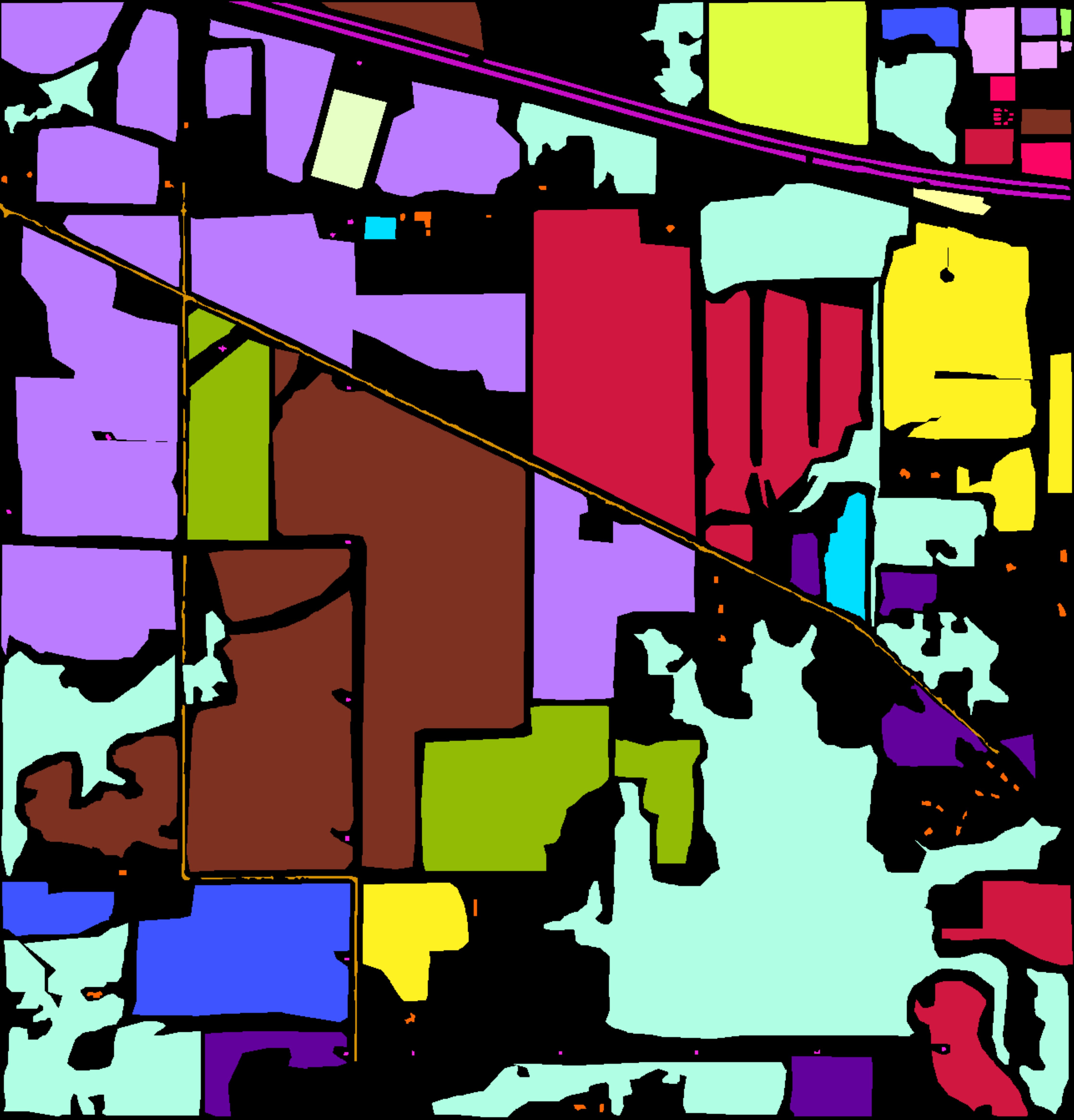} \\
		(a) & (b) \\
	\end{tabular}
	\vspace*{-0.01in}
	\begin{center}
		\includegraphics[width=7.5cm,height=1.7cm]{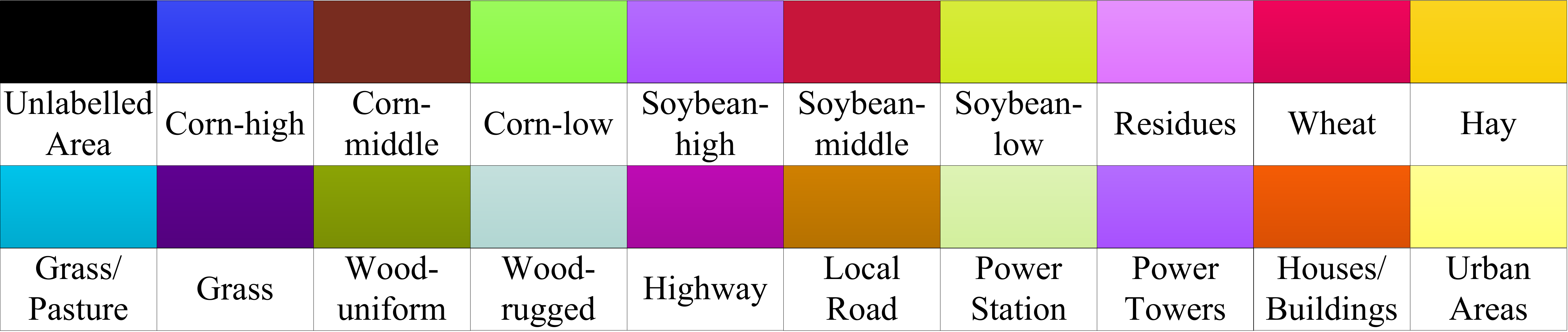}
	\end{center}
	\vspace*{-0.1in}
	\caption{
		(a) True color image and (b) ground-truth of the Indian Pines Data}
	\label{data_i}
\end{figure}

The second hyperspectral data is acquired using ProSpecTIR instrument {in} May 2010 over an agriculture area in Indiana, USA. This image covering agriculture fields has $1342 \times 1287$ spatial dimension with 2$m$ spatial resolution. It has 360 spectral bands over $400 - 2500nm$ wavelength range with approximately $5nm$ spectral resolution. The 19 classes consist of agriculture fields with different residue cover. Fig.~\ref{data_i} shows the true color image of the Indian Pines dataset with corresponding ground truth. 

\begin{algorithm}[htbp]
	\caption{\emph{KcdOLS}}
	\label{alg:kcdols}
	\begin{algorithmic}[1]
		\STATE \textbf{Input:} A training dataset $\boldsymbol{A} = \{\boldsymbol{A}_{l}\}_{l=1}^{c} \in \mathbb{R}^{d \times n}$, where $\boldsymbol{A}_{l}=\{\boldsymbol{a}_{li}\}_{i=1}^{n_{l}} \in \mathbb{R}^{d \times n_l}$, test sample $\boldsymbol{x} \in \mathbb{R}^{d}$, kernel function $\kappa$, sparsity level $S$.
		\\\hrulefill
		\FORALL{$l \in {1,2, \ldots, c}$}
		\STATE Calculate $l$-th class kernel matrix $\boldsymbol{K}_{l} \in \mathbb{R}^{n_{l} \times n_{l}}$ whose $(i,j)$-th entry is $\kappa(\boldsymbol{a}_{li},\boldsymbol{a}_{lj})$ and $\boldsymbol{k}_{l} \in \mathbb{R}^{n_{l}}$ whose $i$-th entry is $\kappa(\boldsymbol{x}, \boldsymbol{a}_{li})$. {Set index set $\Lambda^{1}$ to be the index corresponding to the largest entry in $\boldsymbol{k}_{l}$ and iteration counter $m = 2$}. 
		\WHILE{$m \le S$}
		\STATE \vspace{0.1cm}
		Update the support set $\Lambda^{m} = \Lambda^{m-1} \cup \lambda^{m}$ by solving 					
		\begin{align} \nonumber
			\lambda^{m} &= \argmin_{j \in {1,2,\ldots, n}}(\kappa(\boldsymbol{x},\boldsymbol{x}) \ - \ 2(\boldsymbol{k}_l^{\top})_{\Lambda^{m\!-\!1} \cup j}\tilde{\boldsymbol{\beta}} \ + \ \\
			& \qquad \tilde{\boldsymbol{\beta}}^{\top}(\boldsymbol{K}_l)_{\Lambda^{m\!-\!1} \cup j,\Lambda^{m\!-\!1} \cup j}\tilde{\boldsymbol{\beta}}), \nonumber
		\end{align}
		where {$\tilde{\boldsymbol{\beta}}=((\boldsymbol{K}_l)_{\Lambda^{m\!-\!1} \cup j,\Lambda^{m\!-\!1} \cup j})^{-1}(\boldsymbol{k}_l)_{\Lambda^{m\!-\!1} \cup j}$}. \nonumber	
		\STATE \vspace{0.1cm}
		$m \leftarrow m + 1$
		\vspace{0.1cm}
		\ENDWHILE
		\STATE
		The $l$-th class residual norm can be calculated via
		\begin{equation}
			\nu_l = \sqrt{\kappa(\boldsymbol{y},\boldsymbol{y})-2(\hat{\boldsymbol{\beta}})^{\top}(\boldsymbol{k}_l)_{\Lambda^{m\!-\!1}}+(\hat{\boldsymbol{\beta}})^{\top}(\boldsymbol{K}_l)_{\Lambda^{m\!-\!1},\Lambda^{m\!-\!1}}\hat{\boldsymbol{\beta}}}, \nonumber
		\end{equation}	
		where $\hat{\boldsymbol{\beta}}=\big((\boldsymbol{K}_l)_{\Lambda^{m},\Lambda^{m}}\big)^{\!-1}(\boldsymbol{k}_l)_{\Lambda^{m}}.$
		\ENDFOR
		\STATE Class label of $\boldsymbol{x}$: 
		$\displaystyle{\omega = \argmin_{l = 1,2,\ldots, c}\nu_l}$.
		\\\hrulefill
		\STATE \textbf{Output:} A class label $\omega$.  	  
	\end{algorithmic}
\end{algorithm}

\subsection{Results and analysis}
\label{sec:experiment}
To evaluate the classification performance of cdOLS and KcdOLS, several baseline approaches including SRC, kernel SRC (KSRC), class-dependent OMP (cdOMP), kernel cdOMP (KcdOMP), and linear and nonlinear support vector machine (SVM) are compared. For SRC (KSRC), we use OMP (KOMP) as the recovery method for fair comparison, although convex optimization-based approaches generally outperform greedy-based approaches. Additionally, we also implement the COLS in a class-wise manner (cdCOLS) as well as its kernel version KcdCOLS --- these COLS based variants can be considered as upper bounds in performance of OLS based methods. We also include cdSRC-l1 as a baseline method. It is a class-dependent version of SRC with l1-norm as the constraint, analogous to a class-dependent basis pursuit problem. The kernel functions used in these kernel-based methods was the radial basis function (RBF). The optimal parameters including sparsity level and kernel parameter in RBF are determined via cross-validation. 

The classification results for these two datasets are presented in Table~\ref{tab:result_h} and Table~\ref{tab:result_i} respectively. As expected, we observe that the higher the reconstruction accuracy, the better the classification result. Since COLS is a combinatorial searching method, it is practically unfeasible, particularly when the dictionary size is large. We add it as a comparative method in this work in order to compare the performance gap between cdOLS and cdCOLS. We note that cdCOLS may be feasible in scenarios where the dictionary size is small, and so is the underlying sparsity level for the representations. The overall performance of cdCOLS and cdOLS are similar with a slightly better performance for cdCOLS (as expected). The average performance of cdOLS is generally better than cdOMP.

\begin{table}[htbp]
	\centering
	\caption{Classification accuracy (\%) and standard deviation (in bracket) as a function of training sample size per class for University of Houston data.}
	\resizebox{8.5cm}{!}{
		\begin{tabular}{crrc}
			\toprule
			\textit{\textbf{Algorithm / Sample Size}} & \multicolumn{1}{c}{\textit{\textbf{10}}} & \multicolumn{1}{c}{\textit{\textbf{30}}} & \textit{\textbf{50}} \\
			\midrule
			\textit{\textbf{KcdCOLS}} & \multicolumn{1}{c}{85.8 (1.7)} & \multicolumn{1}{c}{95.2 (0.6)} & 97.3 (0.2) \\
			\textit{\textbf{cdCOLS}} & \multicolumn{1}{c}{84.6 (1.3)} & \multicolumn{1}{c}{93.4 (0.6)} & 96.3 (0.4) \\
			\textit{\textbf{KcdOLS}} & \multicolumn{1}{c}{85.7 (1.7)} & \multicolumn{1}{c}{94.8 (0.7)} & 97.2 (0.3) \\
			\textit{\textbf{cdOLS}} & \multicolumn{1}{c}{84.5 (1.5)} & \multicolumn{1}{c}{92.9 (0.9)} & 95.9 (0.6) \\
			\textit{\textbf{KcdOMP}} & \multicolumn{1}{c}{82.4 (1.3)} & \multicolumn{1}{c}{89.6 (0.8)} & 92.5 (0.5) \\
			\textit{\textbf{cdOMP}} & \multicolumn{1}{c}{79.7 (1.2)} & 87.1 (0.6) & 91.8 (0.5) \\
			\textit{\textbf{KSRC}} & \multicolumn{1}{c}{80.0 (0.9)} & 87.6 (0.7) & 91.8 (0.6) \\
			\textit{\textbf{SRC}} & 78.7 (0.8) & \multicolumn{1}{c}{88.5 (0.5)} & 92.2 (0.6) \\
			\textit{\textbf{cdSRC-l1}} & 77.4 (1.0) & \multicolumn{1}{c}{85.9 (1.3)} & 89.4 (0.8) \\
			\textit{\textbf{SVM-linear}} & 52.8 (2.6) & 55.0 (3.4) & 54.6 (3.4) \\
			\textit{\textbf{SVM-rbf}} & 79.1 (1.2) & \multicolumn{1}{c}{88.8 (0.7)} & 92.9 (0.8) \\
			\bottomrule
		\end{tabular}%
	}
	\label{tab:result_h}%
\end{table}%

\begin{table}[htbp]
	\centering
	\caption{Classification accuracy (\%) and standard deviation (in bracket) as a function of training sample size per class for Indian Pines data.}
	\resizebox{8.5cm}{!}{
	\begin{tabular}{cccc}
		\toprule
		\textit{\textbf{Algorithm / Sample size}} & \textit{\textbf{10}} & \textit{\textbf{30}} & \textit{\textbf{50}} \\
		\midrule
		\textit{\textbf{KcdCOLS}} & 80.2 (1.5) & 89.9 (0.8) & 92.2 (0.5) \\
		\textit{\textbf{cdCOLS}} & 79.0 (1.3) & 88.9 (0.9) & 91.4 (0.5) \\
		\textit{\textbf{KcdOLS}} & 79.1 (1.4) & 86.4 (0.6) & 88.8 (0.8) \\
		\textit{\textbf{cdOLS}} & 78.5 (1.2) & 87.0 (0.7) & 89.4 (0.4) \\
		\textit{\textbf{KcdOMP}} & 78.4 (1.1) & 84.1 (0.9) & 85.3 (0.9) \\
		\textit{\textbf{cdOMP}} & 78.0 (1.2) & 85.3 (0.5) & 86.5 (0.7) \\
		\textit{\textbf{KSRC}} & 65.0 (1.0) & 75.0 (0.8) & 77.8 (1.0) \\
		\textit{\textbf{SRC}} & 69.0 (1.1) & 78.2 (0.6) & 81.0 (1.0) \\
		\textit{\textbf{cdSRC-l1}} & 72.8 (1.4) & 82.7 (1.3) & 86.0 (1.1) \\
		\textit{\textbf{SVM-linear}} & 58.3 (2.6) & 66.6 (1.0) & 67.5 (2.3) \\
		\textit{\textbf{SVM-rbf}} & 71.4 (1.3) & 83.0 (1.6) & 86.5 (0.8) \\
		\bottomrule
	\end{tabular}%
	}	
	\label{tab:result_i}%
\end{table}%

\begin{figure}[htbp]
	\centering
	\includegraphics[width=9cm]{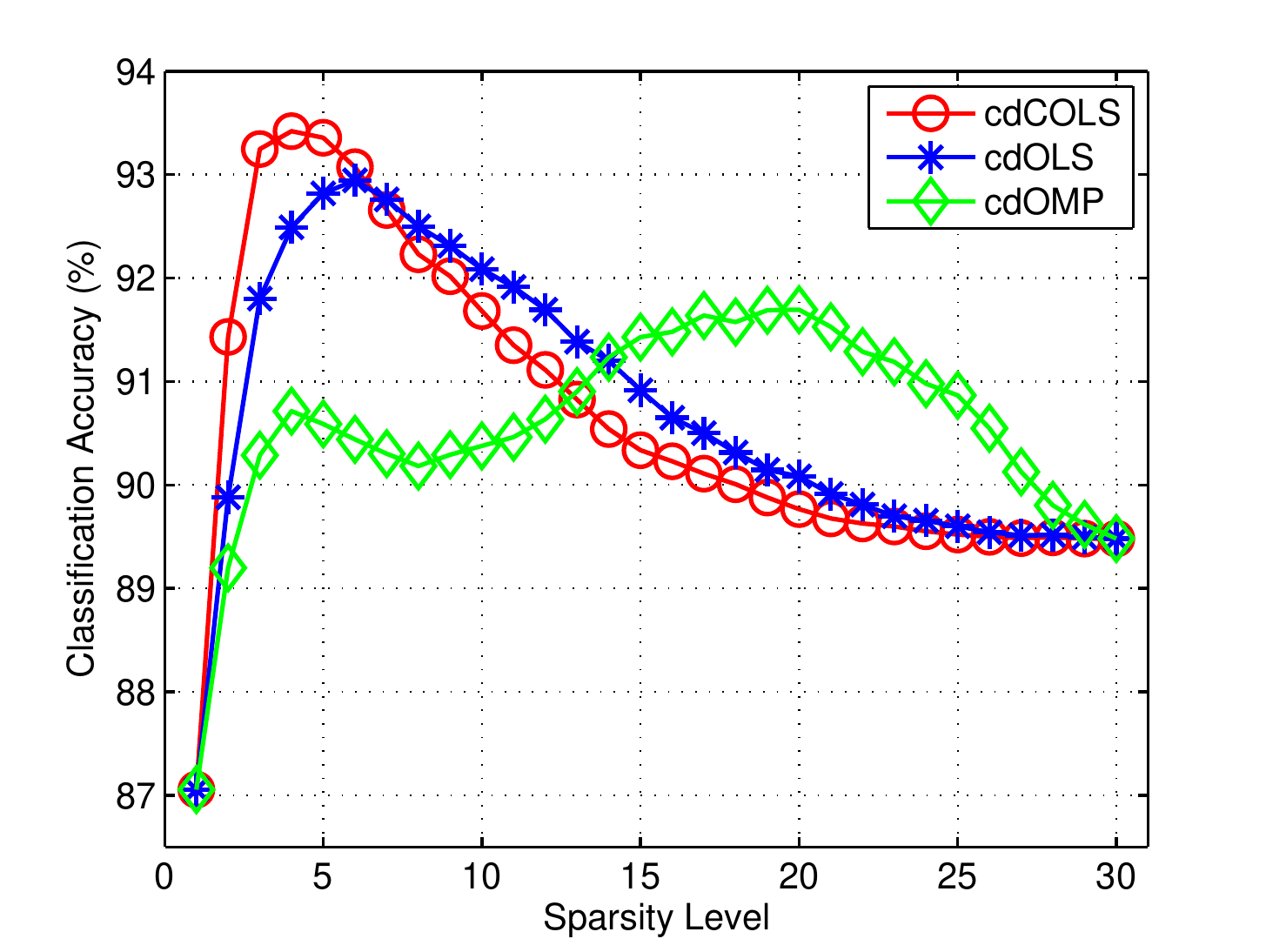} \\
	\caption{Overall classification accuracy (\%) versus sparsity level $S$ for the University of Houston data.}
	\label{fig:spa_h}
\end{figure}

To analyze the effect of sparsity level, we evaluate the performance of cdCOLS, cdOLS and cdOMP under the different sparsity levels. Fig.~\ref{fig:spa_h} show the classification accuracy as a function of sparsity level for University of Houston data respectively. The number of samples per class in this experiment is set to 30. Hence we test the sparsity level starting from 1 to the highest possible number 30. From these two figures, we notice that the optimal sparsity level for these methods are generally very low. This is due to the fact that the within-class hyperspectral data samples are very correlated with each other, and a low residual norm can be derived using a small number of atoms.

Next, we analyze the class-specific residuals obtained for cdCOLS, cdOLS and cdOMP. In this experiment, we select a test sample from class-1 and calculate the residual of the test sample using the training samples from class-1 for both datasets. This experiment is repeated 100 times and the average residuals are reported. Fig.~\ref{fig:res_h} show the residual plots for University of Houston data. As can be seen from the figures, the residual obtained from cdOLS in each iteration is smaller than the residual obtained from cdOMP. Also, the residual obtained from cdOLS is close to the optimal one obtained from cdCOLS in each iteration.

Finally, in order to validate the generalization capabilities of these classifiers, we plot for the University of Houston dataset in Fig.~\ref{fig:map_h} respectively. In this experiment, 30 training samples per class are used. As can be seen from these maps, cdCOLS and cdOLS generally gives much more accurate classification maps compared with cdOMP, especially in the areas of clouds.

\begin{figure}[htbp]
	\centering
	\includegraphics[width=9cm]{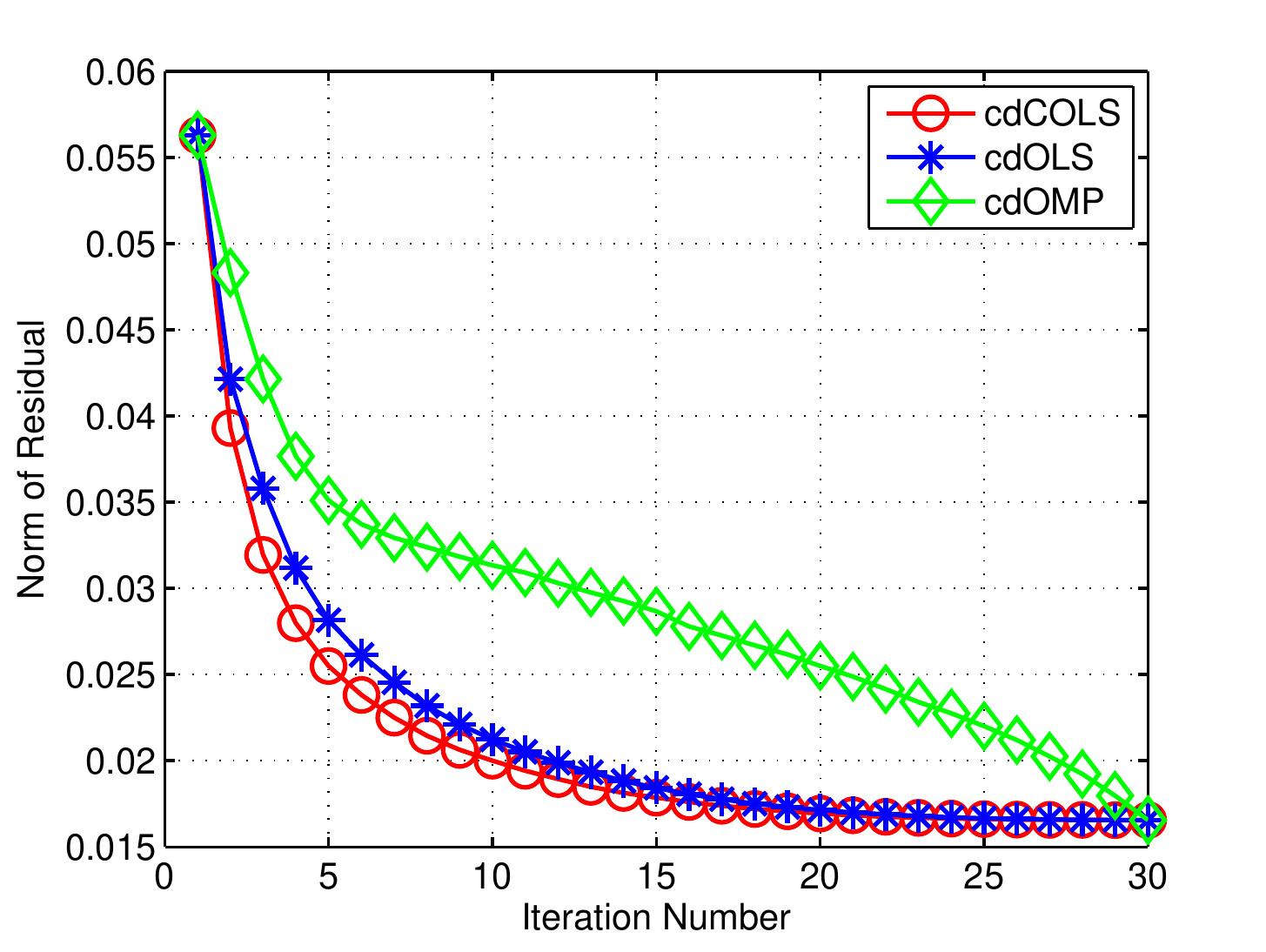} \\
	\caption{Norm of residual versus iteration number for the University of Houston data.}
	\label{fig:res_u}
\end{figure}

\begin{figure}[htbp]
	\centering
	\includegraphics[width=9cm]{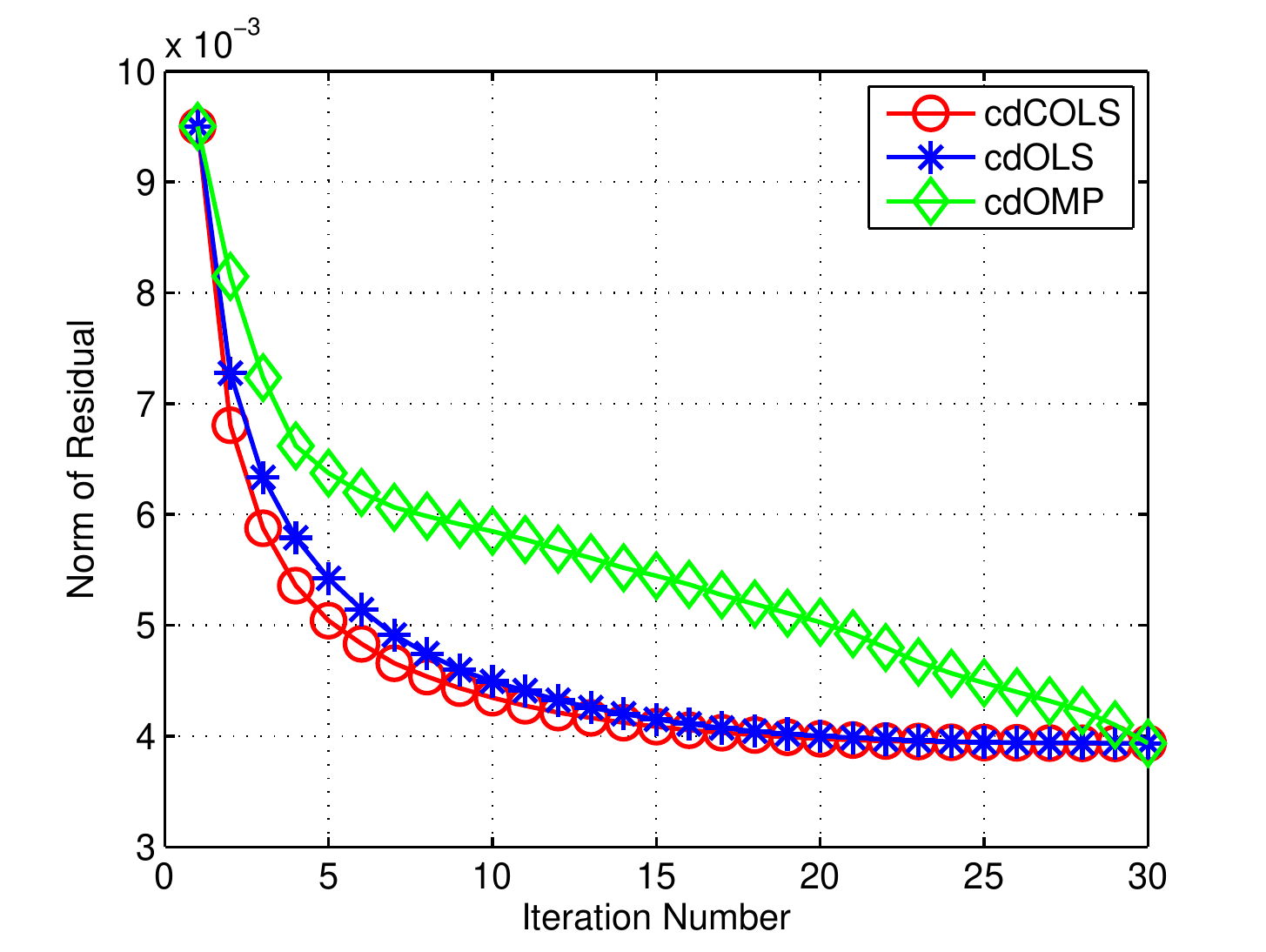} \\
	\caption{Norm of residual versus iteration number for the Indian Pines data.}
	\label{fig:res_h}
\end{figure}

\begin{figure*}[htbp] \small
	\centering
	\begin{tabular}{c}
		\includegraphics[width=14cm]{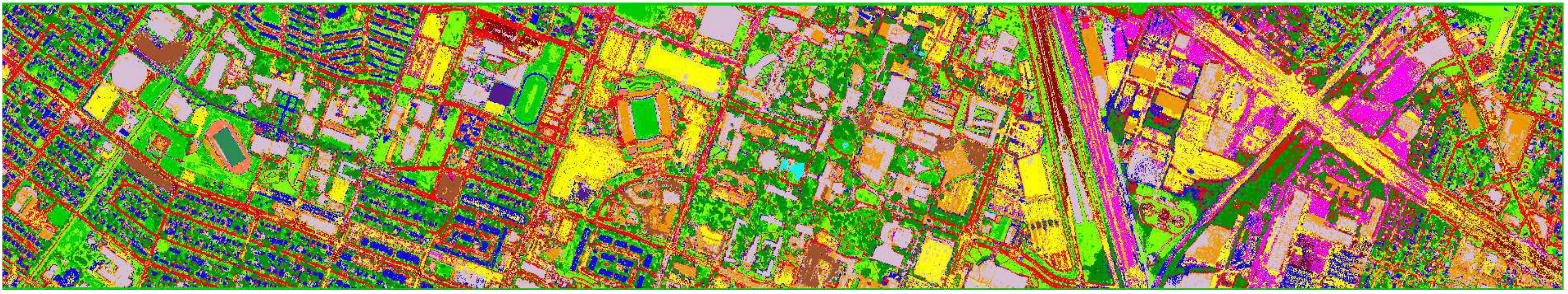} \vspace*{-0.1in} \\
		(a) \\
		\includegraphics[width=14cm]{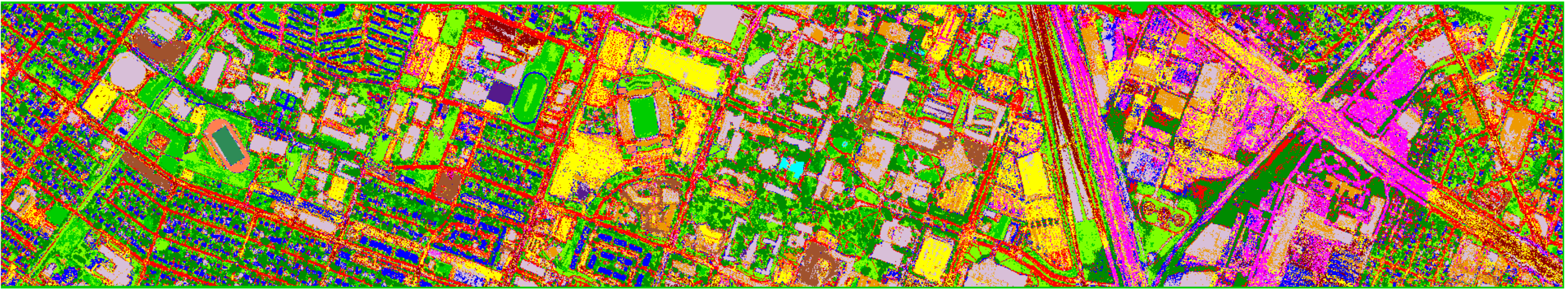} \vspace*{-0.1in} \\
		(b) \\
		\includegraphics[width=14cm]{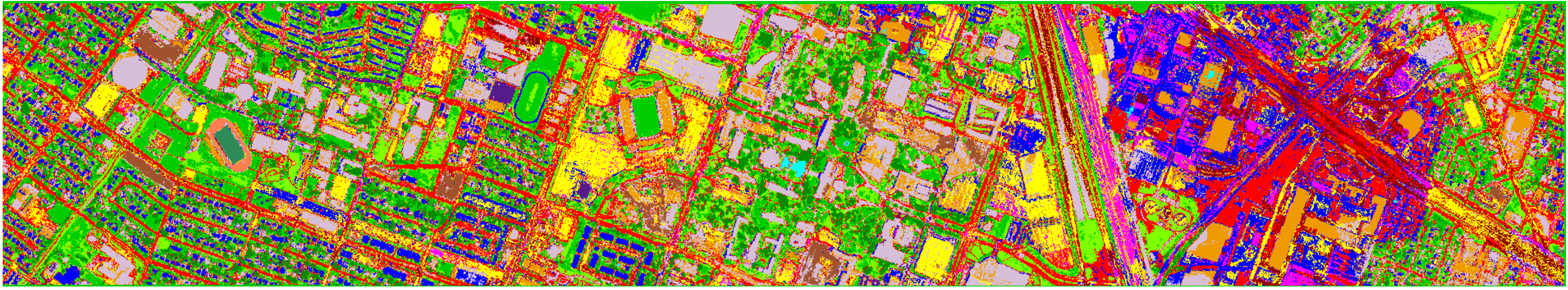} \vspace*{-0.1in} \\
		(c) \\
	\end{tabular}
	\vspace*{-0.1in}
	\begin{center}
		\includegraphics[width=14cm]{figure3_eps-eps-converted-to.pdf} \\
	\end{center}
	\vspace*{-0.1in}
	\caption{Classification maps of University of Houston dataset generated using (a) cdCOLS (b) cdOLS (c) cdOMP.}
	\label{fig:map_h}
\end{figure*}

\section{Conclusion}
\label{sec:conclusion}
In this paper, we present a class-dependent OLS-based classification method named cdOLS for the problem of hyperspectral image classification. We also extend cdOLS into its kernel variant. Through two real-world hyperspectral datasets, we demonstrate that our proposed methods outperform cdOMP, KcdOMP as well as SVM. We also demonstrate that the classification performance of the proposed methods are close to that of cdCOLS and KcdCOLS. Our proposed developments are based on the observation that OLS is generally better suited for sparse coefficient recovery. We also present an \emph{combinatorial} OLS based classifier - COLS, that acts as an upper bound on the performance of such classifiers, and can itself be used as well when the training dictionary is small. For scenarios where training dictionaries are not small, the more feasible cdOLS method has very similar performance to cdCOLS (in both the input and kernel induced space). 

\vspace{-12pt}
\section*{Acknowledgement}
{This work was funded in part by NASA grant NNX14AI47G}.
\vspace{-12pt}


\balance
\bibliographystyle{model2-names}

\end{document}